\documentclass[letterpaper, 10 pt, conference]{ieeeconf}

\IEEEoverridecommandlockouts
\usepackage{graphicx}       
\usepackage{booktabs}       
\usepackage{amsmath}        
\usepackage{gensymb}
\usepackage{microtype}      
\usepackage[utf8]{inputenc} 
\usepackage{newtxmath}      
\usepackage[acronym,shortcuts]{glossaries} 
\usepackage{hyperref}       

\usepackage{flushend}

\usepackage{color}
\usepackage{siunitx}

\usepackage{gensymb}		
\usepackage{subcaption}		

\newcommand{\s}{\operatorname{s}}
\renewcommand{\c}{\operatorname{c}}

\renewcommand{\vec}[1]{\boldsymbol{#1}}   

\newacronym{aov}{AOV}{angle of view}
\newacronym{asctec}{AscTec}{Ascending Technologies}
\newacronym{cpp}{CPP}{coverage path planning}
\newacronym{dof}{DoF}{degrees of freedom}
\newacronym{dz}{DZ}{dropping zone}
\newacronym{ekf}{EKF}{extended Kalman filter}
\newacronym{enu}{ENU}{east north up}
\newacronym{epm}{EPM}{electro permanent magnet}
\newacronym{eurathlon}{euRathlon}{European Robotics League}
\newacronym{euroc}{EuRoC}{European Robotics Challenges}
\newacronym{fsm}{FSM}{finite state machine}
\newacronym{imu}{IMU}{inertial measurement unit}
\newacronym{ipp}{IPP}{informative path planning}
\newacronym{fov}{FOV}{field of view}
\newacronym{lz}{LZ}{landing zone}
\newacronym{mav}{MAV}{micro aerial vehicle}
\newacronym{mbzirc}{MBZIRC}{Mohamed Bin Zayed International Robotics Challenge 2017}
\newacronym{mcu}{MCU}{microcontroller unit}
\newacronym{nmpc}{NMPC}{nonlinear model predictive position control}
\newacronym{msf}{MSF}{Multi Sensor Fusion}
\newacronym{pcb}{PCB}{printed circuit board}
\newacronym{pwm}{PWM}{pulse-width modulation}
\newacronym{sar}{SAR}{search and rescue}
\newacronym{sdk}{SDK}{software development kit}
\newacronym{vio}{VIO}{visual-inertial odometry}
\newacronym{vs}{VS}{visual servoing}

\title{\LARGE\bf Voliro: An Omnidirectional Hexacopter With Tiltable Rotors}

\author{\authorblockN{
	Mina Kamel\authorrefmark{1}\authorrefmark{2},
	Sebastian Verling\authorrefmark{1}\authorrefmark{2},
	Omar Elkhatib\authorrefmark{2},
	Christian Sprecher\authorrefmark{2},\\
	Paula Wulkop\authorrefmark{2},
	Zachary Taylor\authorrefmark{2},
	Roland Siegwart\authorrefmark{2}, and
	Igor Gilitschenski\authorrefmark{3}}
\authorblockA{\authorrefmark{1}shared first authorship,
	\authorrefmark{2}Autonomous Systems Lab, ETH Z\"urich, 
	\authorrefmark{3}CSAIL, MIT, \\
	}
\thanks{Contact: 
mina.kamel@mavt.ethz.ch,
sebastian.verling@mavt.ethz.ch,
ztaylor@ethz.ch,
rsiegwart@ethz.ch,
igilitschenski@mit.edu}
}

\begin{document}
\maketitle
\thispagestyle{empty}
\pagestyle{empty}
\begin{abstract}
Extending the maneuverability of unmanned areal vehicles promises to yield a considerable increase in the areas in which these systems can be used.
Some such applications are the performance of more complicated inspection tasks and the generation of complex uninterrupted movements of an attached camera.
In this paper we address this challenge by presenting Voliro, a novel aerial platform that combines the advantages of existing multi-rotor systems with the agility of omnidirectionally controllable platforms.
We propose the use of a hexacopter with tiltable rotors allowing the system to decouple the control of position and orientation.
The contributions of this work involve the mechanical design as well as a controller with the corresponding allocation scheme.
This work also discusses the design challenges involved when turning the concept of a hexacopter with tiltable rotors into an actual prototype. 
The agility of the system is demonstrated and evaluated in real-world experiments.
\end{abstract}

\section{Introduction} \label{sec:introduction}
%
%
As autonomous unmanned aerial vehicles (UAVs) are being deployed in a wide area of real-world robotics applications~\cite{bircher2016receding,gawel2017aerial,treecavity2016,BorderPatroUAV1}, there is a growing demand to broaden the scope of their applicability, effectiveness, and robustness.
Due to this demand, improving a UAV's agility and enabling aerial manipulation has received considerable interest.
Advances in these areas may unlock deployment in numerous use-cases.
For instance, drone-based inspection of bridges~\cite{bircher2016three} may require the drone to collect measurements with a sensor that must remain in contact with the bridge structure at all time.
This requires the drone to adapt its own orientation to the bridge's geometry, while controlling its position independently.
Enabling a drone to reach and hold any possible pose also gives rise to many further applications such as enabling uninterrupted complex camera motions for film shots~\cite{nageli2017real}.
These are only two of the many possible applications that would benefit from overcoming the limits imposed by the dynamics of most commercially available multi-rotor UAVs.

%
%
The demand and interest for novel UAV concepts has been mirrored by numerous research projects.
In  \cite{Verling2016}, a small Vertical Take-Off and Landing (VTOL) platform  is proposed.
The system begins its flight as a multi-copter, before transitioning into a fixed-wing mode for comparably long-endurance surveillance and inspection flights.
Handling of strong winds for platforms of this size is investigated in~\cite{Furieri2017}.
This concept comes at the cost of a lower agility compared to a "classical" multi-copter.
The \emph{Omnicopter}~\cite{Brescianini2016} can generate forces and torques in any direction.
This is achieved by intelligent rotor placement in a cube-like structure.
%
It has the downside of counteracting forces that can reduce the systems efficiency and hence its flight time and only a limited usefulness of certain rotors in a given configuration.
%

%
%
The use of tilting rotors promises to address some of the aforementioned challenges.
A concept and simulation of a multi-copter involving tilting rotors was presented in~ \cite{Ryll2012} and experimentally evaluated in~\cite{ryll2013first}.
More approaches were evaluated in~ \cite{Ryll2016,Ryll2017}.
While improving upon maneuverability, these approaches do not offer omnidirectional flight capabilities.
The work in \cite{Oosedo2015} also involved tilting rotors, but only considered two distinct flight modes, one in  horizontal and one in vertical orientation.
The concept presented in~\cite{Kaufman2014} is the most similar design we are aware of to the proposed Voliro platform.
However, this work merely considered control of orientation and did not propose any approach for jointly controlling position and orientation.
While the aforementioned works pushed the boundaries of multi-copter agility, they did not demonstrate omnidirectional flight over the whole flight envelope of a system with tilting rotors.
\begin{figure}[t!]
	\centering
	\includegraphics[width=0.99\linewidth, trim={0 0 500 0},clip]{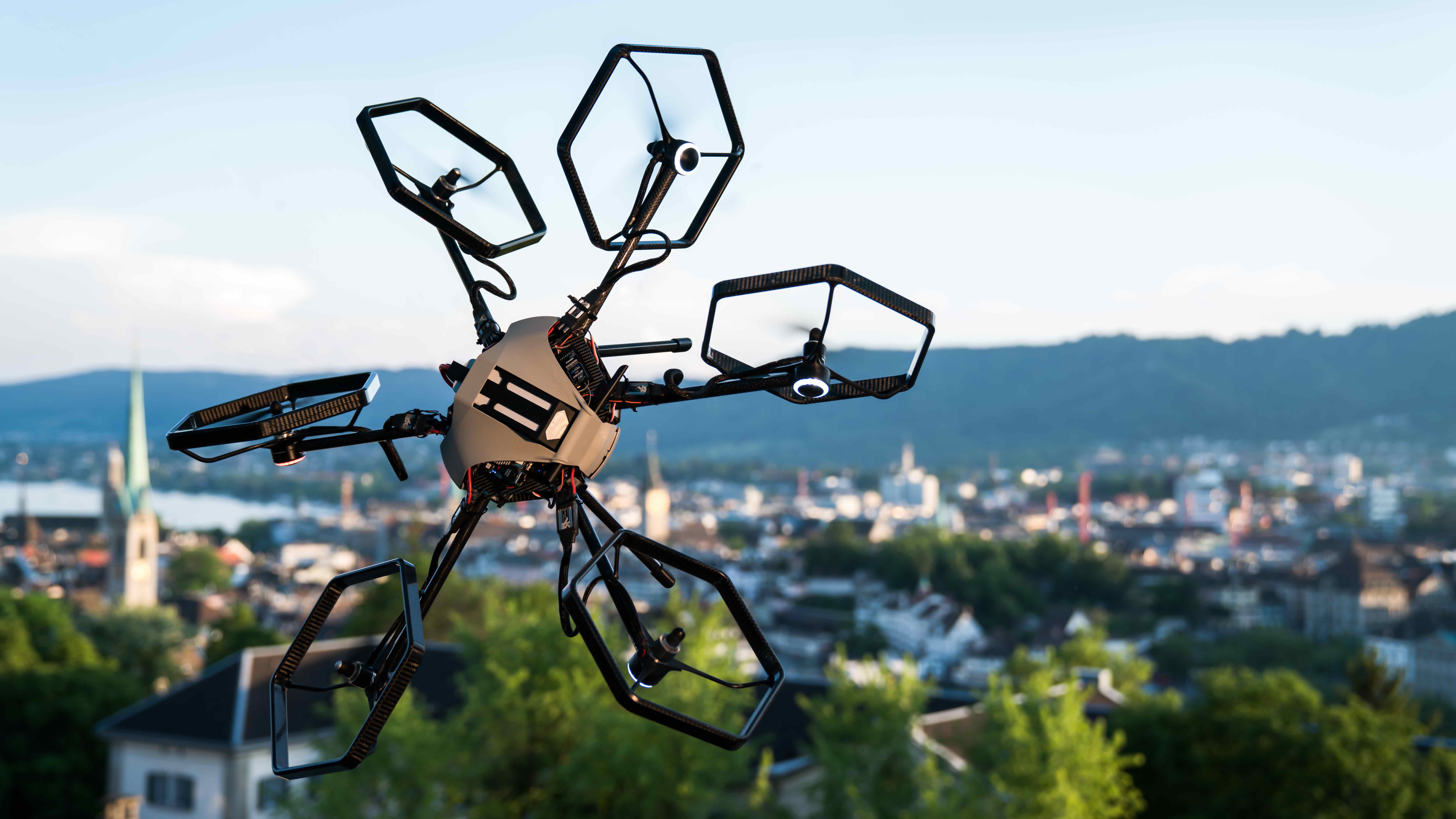}
	\caption{The final version of Voliro platform.}
	\label{fig:colibrigeil}
\end{figure}

%
%
Our goal is to overcome the limitations of common commercially available multi-copters while preserving their advantages.
The main conceptual challenge for achieving this goal is the mechanical design and control of the system.
To achieve this the mechanical design is required to address three challenges:
(1) Incorporating the hardware that enables additional maneuverability while providing a low weight;
(2) Placing the propellers such that they can contribute to the system in any configuration and generate as little counteracting forces as possible;
(3) Safe design such that a tool can be easily mounted and the drone operated close to a given surface.
These three challenges also translate into challenges for designing the controller.

%
%
In this work, we present \emph{Voliro}, a hexacopter with tiltable rotors.
As the rotor orientation can be fully controlled, this system allows for decoupling the control of position and orientation.
It can be easily controlled through a PID controller and a simple allocation scheme which translates the control output into motor configurations.
A prototype system is demonstrated and controlled in a large range of configurations while utilizing a Vicon system to provide state estimation.
The contributions of this work can be summarized as follows:
\begin{itemize}
\item  The mechanical design of a fully-controllable hexacopter with tilting rotors.
\item  The corresponding control and allocation schemes.
\item Evaluation of the concept in real-world experiments on a prototype platform.
\end{itemize}

\section{System Description}
\label{sec:system}
\subsection{Mechanical Design}

\begin{figure}[!htbp]
   \centering
   \includegraphics[width=0.4\textwidth]{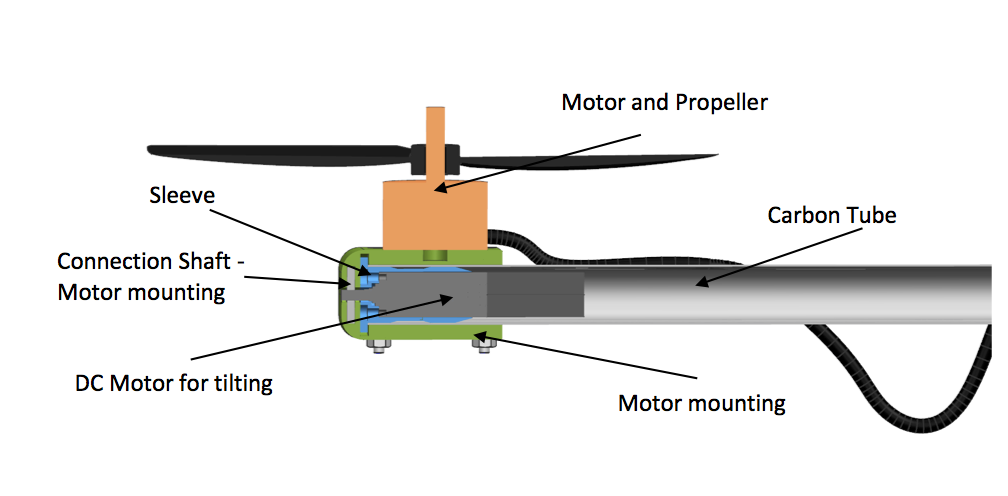}
   \caption{Section view of a rotor unit}
   \label{fig:RotorUnit}
\end{figure}

The prototype presented in this work has six rotors arranged evenly on a circle (Figure \ref{fig:colibrigeil}) in a traditional helicopter configuration.
These rotors arms are made of carbon fiber tubes and each rotor has the ability to rotate around the axis of the arm.
Only the motor and its housing rotates. This rotation is achieved using brushless DC motors, which are inserted at the end of the arms, as can be seen in Figure~\ref{fig:RotorUnit}.
The arms themselves are rigidly attached to the platform main frame. This gives increased structural rigidity of the system when compared to designs that would rotate the entire arm.

The tilting motors are \SI{12}{mm} in diameter and are equipped with three Hall sensors and a high ratio gearbox for precise position estimation and control.
The tilting motors have a maximum rotation speed of \SI{7.85}{rad/s} and a maximum torque of \SI{0.5}{Nm}. This allows for fast control of the tilting angles. The decision to use this motor setup over RC servo motors was made as such servos are typically limited to \ang{180} of rotation, which would limit the flight envelope of the platform. A further advantage of the brushless motors is their narrow profile which allowed them to be hidden inside the platforms tubular arms.
With this setup, the rotors are capable of rotating \ang{720} around their axes. The limiting factor in this design is the winding of the thrust motor cables around the arms. \\




\subsection{Hardware}

%
%
This section introduces the main electronic and mechanical components of the system.
For the generation of thrust 9 inch DJI propellers and KDE2315XF-885 motors were used.
This propultion system is capable of providing a maximum of \SI{13.7}{N} of thrust per motor. 
The system requires a high thrust-to-weight ratio in order to be able to hover at large inclination angles, where most rotors cannot orient their force vector to directly oppose gravity.
Faulhaber 1226S012B1 brushless DC motors were utilized in the rotor tilting system. 
Each of these tilting motors requires a motion controller which measures the current position of the motor shaft and applies the required voltage to control its position.
To power the motors and the electronic components at different voltages, a 6-cell LiPo Battery and accompanying power distribution boards have been used.
Table \ref{tab:components} shows the full list of electronic components on this platform.

%
%
In order to house these electronics, a special core was designed using customized parts.
The core is made out of two \SI{1.5}{mm} thick carbon fiber plates and assembled using aluminum screws.
Aluminum spacers keep the plates a fixed distance apart to provide room for the electronics. They also act to clamp the carbon tubes on which the rotor units are mounted.
Inside of the core are a Pixhawk flight controller, an UP Board embedded computer that provides additional computational power, and the motion control boards for the tilting motors.

%
%
The presented system has a total weight of \SI{3.2}{kg} and a flight time of around \SI{8}{min} in its horizontal orientation. 
This flight time decreases significantly if the system is required to hover at a large inclination angle.

\begin{figure}[!htbp]
\begin{center}
\begin{tabular}{|l|l|}
\hline
\bf Component & \bf Name \\ \hline
Thrust Motors & KDE2315XF-885\\ \hline
ESC & KDEXF-UAS35 \\ \hline
Tilt Motor & 1226S012B1 \\ \hline
Motion control board & MCBL 3002 \\ \hline
Battery & Swaytronic LiPo 6S \\ \hline
PDB & Piko PDB\\ \hline
BEC 12 V & micro BEC \\ \hline
BEC 5 V & Reely UBEC  \\ \hline
Onboard computer & UP Board \\ \hline
Flight controller & Pixhawk  \\ \hline
RC receiver & Futaba R3008SB \\ \hline
WiFi Antenna & Dlink DWA-172 5 GHz \\ \hline
\end{tabular}
\caption{Components and suppliers} \label{tab:components}
\end{center}
\end{figure}

\subsection{Flight Controller}

The Pixhawk flight controller used on this platform consists of an Inertial Measurement Unit (IMU), magnetometer, barometer, and a Cortex-M4F microprocessor.
The Pixhawk runs the PX4 software~\cite{meier2015px4}, which handles full control of the multicopter as well as the interfaces with other devices. 
It also provides a flexible and modular framework that allows the integration of new control schemes.

Using the built-in state estimation provided by the PX4 software, the sensor data obtained by the IMU is fused together with the external pose information from a Vicon motion capture system or GNSS in the case of outdoor flight to provide an estimate for the system's position.
This information enters the controller block together with the desired pose trajectory which is generated by the user on an external computer. 
This desired trajectory is then sent to the UP board via Wifi, and ultimately to the Pixhawk through a serial connection.
The controller block contains both the position and attitude controller, as well as an allocation block that maps the desired forces and moments onto the twelve actuators. This is crucial to the decoupling of position and orientation.
After the desired control inputs are calculated, they are fed into the mixer block, which maps these values to the actuator PWM signals.

\begin{figure}[!htbp]
   \centering
   \includegraphics[width=0.45\textwidth]{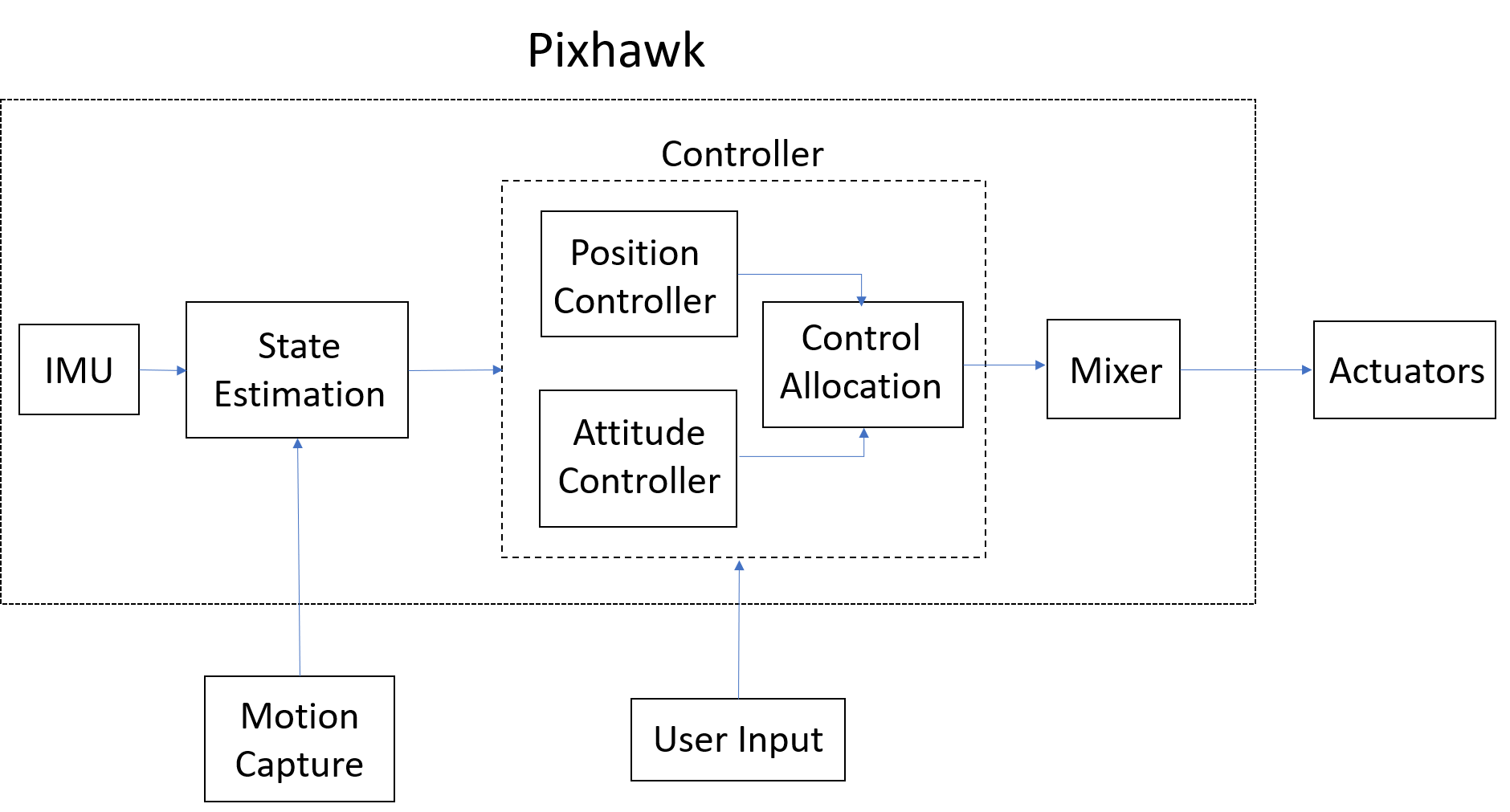}
   \caption{Flight controller structure (A rough prototype)}
   \label{fig:colibri}
\end{figure}

\section{Modelling}
\label{sec:modelling}
The development of a system model for such a vehicle is a necessary step in enabling model-based control synthesis and thorough testing.
To fulfill this task a methodology that combines rigid-body dynamics with well established aerodynamic modeling techniques was used.

\subsection{Coordinates and Conventions}
In this section the coordinate systems and notation conventions used are explained.
\subsubsection{Coordinate Frames}
\begin{figure}
  \centering
  \includegraphics[width=0.8\columnwidth]{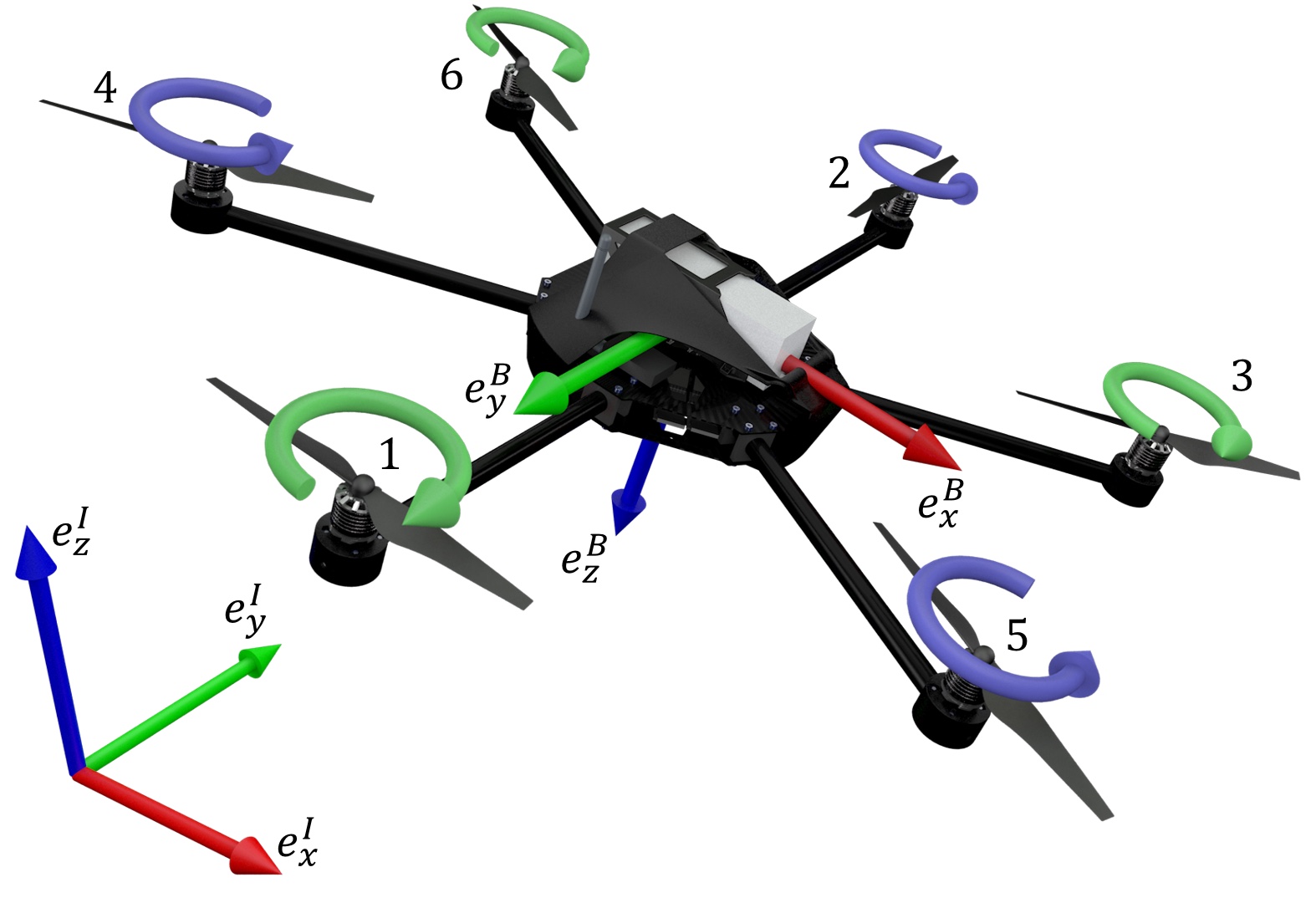}
  \caption{Interial and Body Coordinate System}
  \label{fig:coordinateSystem}
\end{figure}

\begin{figure}
  \centering
  \includegraphics[width=0.8\columnwidth]{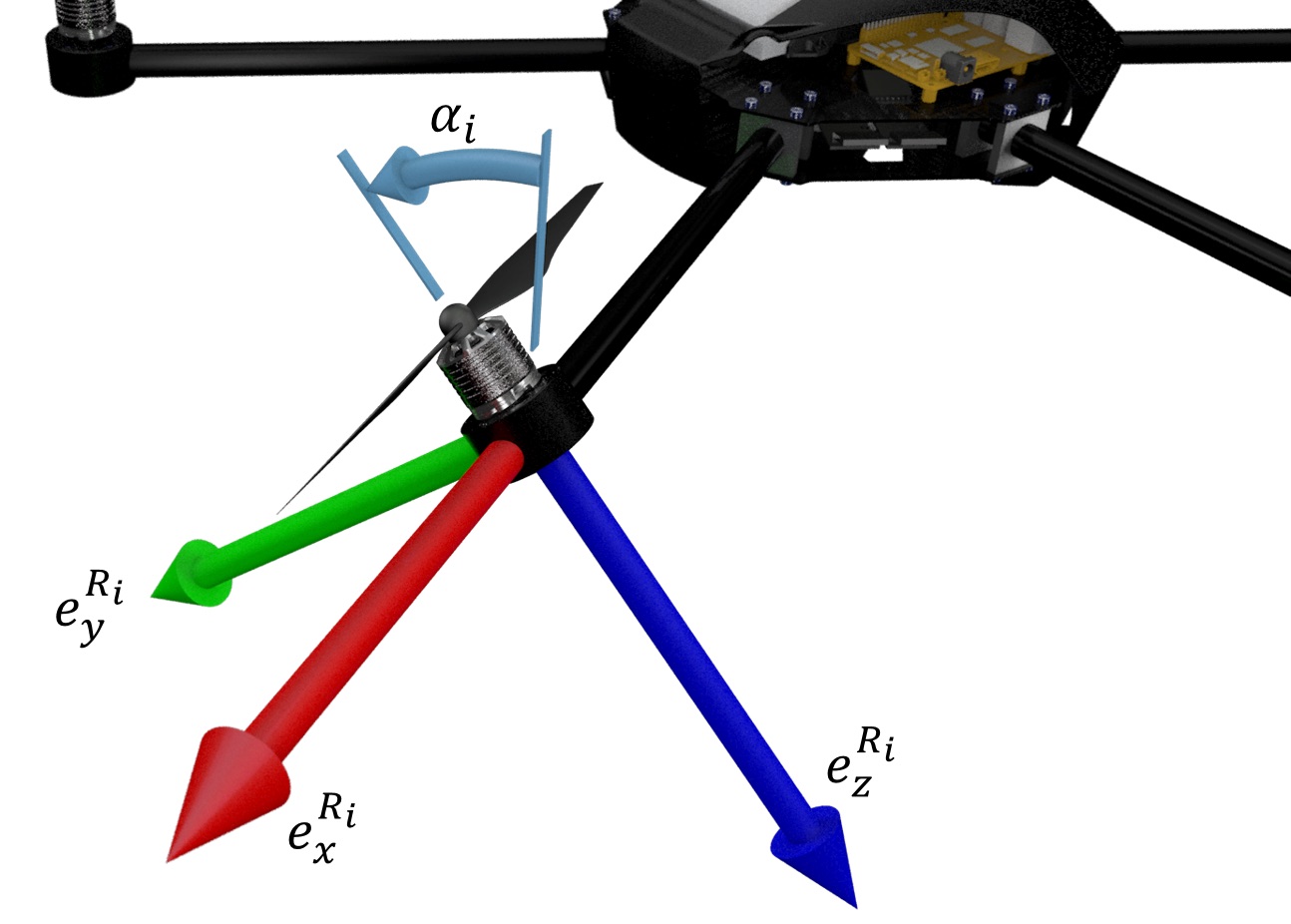}
  \caption{Rotor Coordinate System}
  \label{fig:rotorCoordinateSystem}
\end{figure}

Overall eight coordinate frames were used.
The inertial frame $\mathcal{F}_i$ is fixed on the ground and its z-axis points upwards.
In contrast, the z-axis of the body frame $\mathcal{F}_B$ points downwards as seen in Figure \ref{fig:coordinateSystem}.
Its origin is at the center of gravity.
And finally, there are the six coordinate frames of the rotor units $\mathcal{F}_{R,i}$ shown in Figure~\ref{fig:rotorCoordinateSystem}.
Their origins are at the center of the rotor blades and their x-axes are aligned with the axes to the center of mass of the body.
Theses coordinate frames rotate around their x-axes with respect to the body frame.


\subsubsection{Notation convention}
Throughout this paper we use left-hand subscripts for the coordinate system a symbol is referenced in.
E.g. $_I\boldsymbol{r}$ stands for a vector $\boldsymbol{r}$ represented in coordinate system $\mathcal{F}_I$.
The Matrix $\boldsymbol{R}_{AB}$ stands for the rotation matrix that rotates a vector represented in coordinate system $\mathcal{F}_B$ to the coordinate system $\mathcal{F}_A$
\begin{align*}
{}_A\boldsymbol{r} = \boldsymbol{R}_{AB} \cdot {}_B\boldsymbol{r}
\end{align*}
The estimate of a state $x$ will be denoted with a hat on top of the symbol: $\hat{x}$.

\subsection{Assumptions}
For the sake of model simplicity, the following assumptions have been made:
\begin{itemize}
    \item The body structure is rigid and symmetric.
    \item The rotors are on the same height as the center of gravity and rotate around an axis that passes through the center of gravity.
    \item The rotors are rigid.
    \item Thrust and drag torque are proportional to the square of rotor’s speed.
    \item The linear and angular velocity of the body are small.
    \item The dynamics of the tilting motors are independent of the rotational speed of rotors.
    \item The thrust and drag torque produced by one rotor is independent of the thrust and torque of the other rotors, that is the produced air flows do not interference.
\end{itemize}
While most of these assumptions will cause little or no deviation of the flight characteristics of the real world system from the derived system model, the last assumption which ignores rotor interference is likely to have a significant impact.
However in our experience and flight tests we have found that with a well tuned controller this influence can be treated as an unmodeled disturbance.
\subsection{Rigid Body Model}
For the modeling of the body dynamics, the Newton-Euler formalism was used.
The general form of this formalism is the following:
\begin{align}
    \begin{bmatrix}
        m\boldsymbol{\mathcal{I}_3} & \boldsymbol{0} \\
        \boldsymbol{0} & \boldsymbol{J}
    \end{bmatrix}
    \begin{bmatrix}
        {}_B\boldsymbol{\dot{v}} \\
        {}_B\boldsymbol{\dot{\omega}}
    \end{bmatrix}
    +
    \begin{bmatrix}
        {}_B\boldsymbol{\omega} \times (m\cdot {}_B\boldsymbol{v})\\
        {}_B\boldsymbol{\omega} \times (I\cdot {}_B\boldsymbol{\omega})
    \end{bmatrix}
    =
    \begin{bmatrix}
        {}_B\boldsymbol{F} \\
        {}_B\boldsymbol{M}
    \end{bmatrix} v
\end{align}
with $\boldsymbol{\mathcal{I}_3}$ being the 3-dimensional identity matrix and $\boldsymbol{J}$ the moment of inertia of the system.

\subsection{Motor Dynamics}
The motor dynamics of the DC motors of the rotors are modeled as a first order system.
\begin{align}
    \dot{n}_i = \frac{1}{\tau_{n}}(n_{i,des} - n_i)
\end{align}
with $n_{i,des}$ being the desired angular velocity and $\tau_{n}$ is the motor's time constant.
Similarly, the closed loop tilting velocity $ \omega_{\alpha_i} = \dot{\alpha_i} $ dynamics are modeled as first order system with time constant $ \tau_{\alpha}\ $. 
Moreover, the tilting velocity is saturated to model the maximum possible tilting velocity as follows
\begin{equation}
| \omega_{\alpha_i}  | \leq \omega_{\alpha, max} ,
\end{equation} 
where $ \omega_{\alpha, max} $ is the maximum possible tilting velocity.
%

\subsection{Aerodynamics}
The aerodynamics modeling transfers the actual angular velocities $n$ and tilting angles $\alpha$ into the forces $\boldsymbol{F}$ and moments $\boldsymbol{M}$ acting on the center of gravity of the body.
Given the stated assumptions of low body velocities, we can neglect the aerodynamic drag of the main body and assume that the thrust and reaction torque are proportional to the square of the rotational velocity of the rotors. 
The force and torque produced by one rotor is given by:
\begin{align}\label{eq:rotor_force_torque}
    F &= \mu n^2 \\
    \tau &= \kappa n^2
\end{align}
with $\mu$ being the lift force coefficient and $\kappa$ the drag torque coefficient. 
These forces and moments are produced in the z-direction of the coordinate frames of the rotor.
The forces have a negative sign, since the z-axis is pointing downwards.

The forces and moments are then rotated into the body frame.
\begin{align}
    {}_{R_i}\boldsymbol{F}_i &= -\mu n_i ^{2}  \cdot{}_{R_i}\boldsymbol{e}_z \\
    {}_{R_i}\boldsymbol{\tau}_i &= - c_i \kappa n_i^{2} \cdot{}_{R_i}\boldsymbol{e}_z
\end{align}
with $ c_i
\begin{cases}
    1, & \text{for } i = 1,3,6 \\ 
    -1, & \text{for } i = 2,4,5
\end{cases} $

\begin{align}
     {}_B\boldsymbol{F} &= \sum_i {}_B\boldsymbol{F}_i = \sum_i \boldsymbol{R}_{BR_i}\cdot{}_{R_i}\boldsymbol{F_i} \\
     {}_B\boldsymbol{M} &= \sum_i {}_B\boldsymbol{\tau}_i + \sum_i {}_{R_i}\boldsymbol{M}_{ind,i} = \sum_i \boldsymbol{R}_{BR_i}\cdot{}_{R_i}\boldsymbol{\tau_i} + {}_B\boldsymbol{r}_i \times {}_B\boldsymbol{F}_i
\end{align}
where $\boldsymbol{R}_{BR_i}$ is the rotation matrix of the i-th rotor frame to the body frame and ${}_B\boldsymbol{r}_i$ is the distance from the center of gravity to the i-th rotor.
The relation between the forces, moments, tilting angles and angular velocities can then be described by the following
equation.
\begin{align}\label{eq:allocation_mapping}
    \begin{pmatrix}
        \boldsymbol{F} \\
        \boldsymbol{M}
    \end{pmatrix}
    = \boldsymbol{A}(\boldsymbol{\alpha})\boldsymbol{N}
\end{align}
where $\boldsymbol{A}(\boldsymbol{\alpha}) \in \mathbb{R}^{6\times6}$ (see Equation \eqref{eq:allocation_matrix}), $\boldsymbol{N} = (n_1^{2}, n_2^{2}, ..., n_6^{2})$
and with $\c(\cdot)$ representing cosine and $s(\cdot)$ representing sine.
{
\begin{figure*}[!t]
\tiny

\begin{align}\label{eq:allocation_matrix}
    \boldsymbol{A}(\boldsymbol{\alpha}) =
    \begin{bmatrix}
        -\mu \s(\alpha_1) & \mu \s(\alpha_2) & \mu \frac{1}{2}\s(\alpha_3) & -\mu \frac{1}{2} \s(\alpha_4) & -\mu \frac{1}{2} \s(\alpha_5) & \mu \frac{1}{2} \s(\alpha_6) \\
        0 & 0 & \mu \frac{\sqrt{3}}{2}\s(\alpha_3) & -\mu \frac{\sqrt{3}}{2} \s(\alpha_4) & \mu \frac{\sqrt{3}}{2} \s(\alpha_5) & -\mu \frac{\sqrt{3}}{2} \s(\alpha_6) \\
        -\mu \c(\alpha_1) & -\mu \c(\alpha_2) & -\mu \c(\alpha_3) & -\mu \c(\alpha_4) & -\mu \c(\alpha_5) & -\mu \c(\alpha_6) \\
        -\mu l \c(\alpha_1) - \kappa \s(\alpha_1) & \mu l \c(\alpha_2) - \kappa \s(\alpha_2) & \mu l \frac{1}{2}\c(\alpha_3) + \kappa \frac{1}{2} \s(\alpha_3) & -\mu l \frac{1}{2}\c(\alpha_4) + \kappa \frac{1}{2} \s(\alpha_4) & -\mu l \frac{1}{2}\c(\alpha_5) + \kappa \frac{1}{2} \s(\alpha_5) & \mu l \frac{1}{2}\c(\alpha_6) + \kappa \frac{1}{2} \s(\alpha_6) \\
        0 & 0 & \mu l \frac{\sqrt{3}}{2}\c(\alpha_3) + \kappa \frac{\sqrt{3}}{2} \s(\alpha_3) & -\mu l \frac{\sqrt{3}}{2}\c(\alpha_4) + \kappa \frac{\sqrt{3}}{2} \s(\alpha_4) & \mu l \frac{\sqrt{3}}{2}\c(\alpha_5) - \kappa \frac{\sqrt{3}}{2} \s(\alpha_5) & -\mu l \frac{\sqrt{3}}{2}\c(\alpha_6) - \kappa \frac{\sqrt{3}}{2} \s(\alpha_6) \\
        \mu l \c(\alpha_1) - \kappa \s(\alpha_1) & \mu l \c(\alpha_2) + \kappa \s(\alpha_2) & \mu l \frac{1}{2}\c(\alpha_3) - \kappa \frac{1}{2} \s(\alpha_3) & \mu l \frac{1}{2}\c(\alpha_4) + \kappa \frac{1}{2} \s(\alpha_4) & \mu l \frac{1}{2}\c(\alpha_5) + \kappa \frac{1}{2} \s(\alpha_5) & \mu l \frac{1}{2}\c(\alpha_6) - \kappa \frac{1}{2} \s(\alpha_6) \\
    \end{bmatrix}
\end{align}
\end{figure*}
}
$\boldsymbol{A}(\boldsymbol{\alpha})$ will henceforth be referred to as the allocation matrix.
For a conventional quad- or hexacopter this allocation matrix is static.
However, for this project it is a function of the tilting angles $\alpha$, which is an advantage, since the total force vector can be directed in any orientation through the selection of appropriate tilting angles.
In addition to this, the matrix is of size $6\times6$, as opposed to the $4\times6$ size utilized by a conventional hexacopter. 
This means the system can produce a desired force in all three axes, instead of just the z-direction.
Since the matrix has full rank in most configurations (it’s singular in some special cases such as when all $\alpha$'s are zero and other more complex cases), all forces and moments can be produced independently, meaning that  translation and  rotation of the platform can be controlled independently.

\section{Control}
\label{sc:control}
\subsection{Control structure}
\begin{figure}
	\centering
	\includegraphics[width=0.5\textwidth]{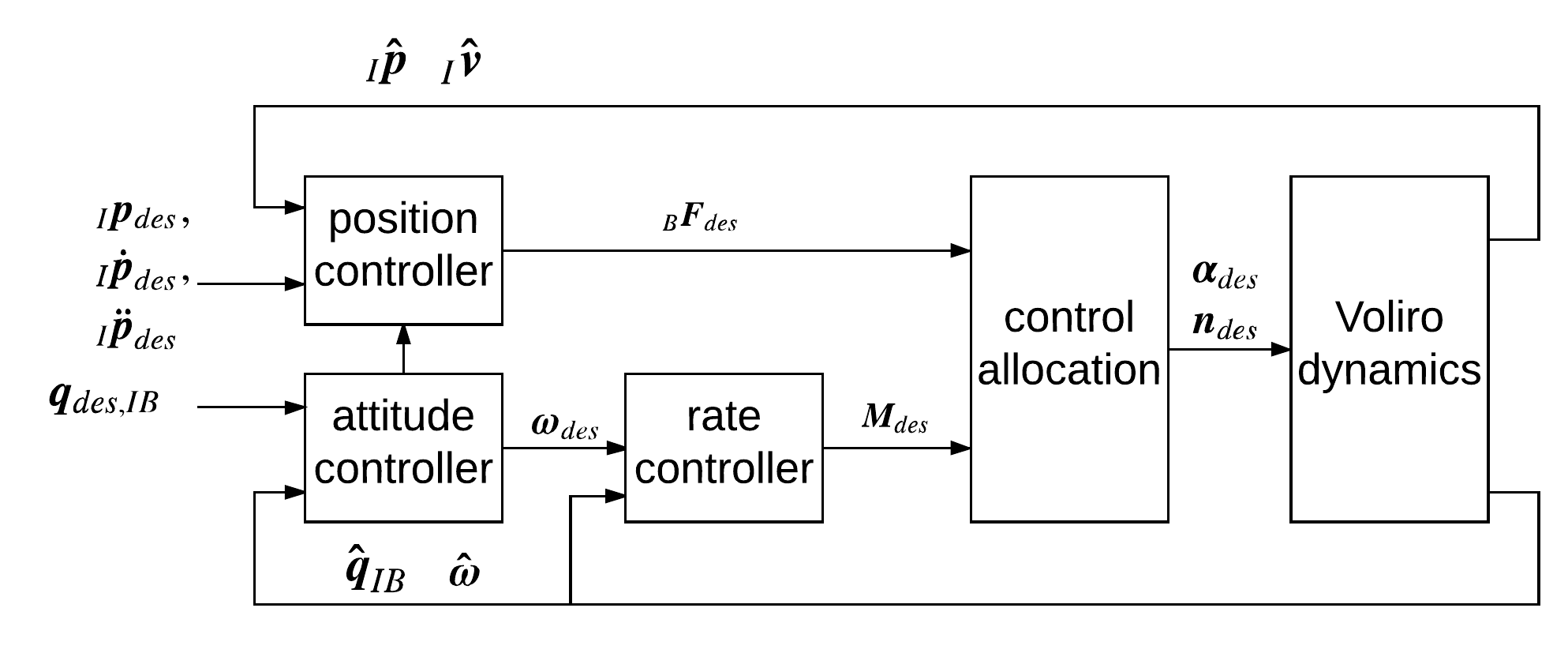}
	\caption{Control structure of Voliro platform. The controller is running onboard the PixHawk flight
		controller.}
	\label{fig:control_block_diagram}
\end{figure}

In this section, we present a baseline controller for the Voliro platform.
Given the omni-directional  nature of the platform, it is possible to decouple the position and attitude dynamics as the actuation forces and torques are independent.
Therefore, we consider two separate controllers for position and  attitude reference tracking. A cascade control structure is employed to control the attitude as shown in Figure~\ref{fig:control_block_diagram}.

Desired forces and torques are achieved by changing the angular velocity and the orientation of each propeller.
Therefore, the control allocation problem becomes challenging, as for the 6 \ac{dof} output, there are $12$ control inputs.
We  introduce the position and attitude controllers, followed by a novel solution to this control allocation problem.

\subsection{Position control}
The position control is based on a PID controller with a feed forward terms that generates a desired force vector in the vehicle body frame $ \mathcal{F}_B $.
The position error is defined as the difference between desired and estimated position $ {}_{I}\boldsymbol{p}_{err} =
{}_{I}\boldsymbol{p}_{des} - {}_{I}\boldsymbol{\hat{p}} $.
The position control law is given by:

\begin{equation}\label{eq:position_controller}
	\begin{split}
	{}_{B}\boldsymbol{F}_{des} = \boldsymbol{R}_{IB}^{T} \left( k_{p,p} {}_{I}\boldsymbol{p}_{err}  + k_{d,p}{}_{I}\boldsymbol{\dot{p}}_{err} + \vphantom{\int_1^2} \right. \\
	\left. k_{i, p}\int{\boldsymbol{p}_{err}dt} + m\boldsymbol{g} + {}_{I}\boldsymbol{\ddot{p}}_{des} \right)
\end{split}
\end{equation}
where $ \boldsymbol{R}_{IB} $ is the rotation matrix between the body frame $ \mathcal{F}_B $ and inertial frame, $ \mathcal{F}_I $, $ k_{p,p}, k_{d, p},  k_{i, p}$ are the proportional, derivative and integral gains respectively, $ m $ is the vehicle mass and $ \boldsymbol{g} $ is the gravity vector.

\subsection{Attitude control}
The attitude controller consists of two cascade controllers, an outer loop for the attitude control that generated desired body rates and a rate controller that generates desired moments.
The attitude controller is based on quaternion error which is given by:
\begin{equation}
\boldsymbol{q}_{err} = \boldsymbol{q}_{des, IB} \otimes \boldsymbol{\hat{q}}_{IB}^{*} = \left( \begin{array}{c}
q_{w, err} \\
\boldsymbol{q}_{v, err}
\end{array} \right)
\end{equation}
The desired body rate $ \boldsymbol{\omega}_{des} $ is generated from the vector part of the quaternion error $ \boldsymbol{q}_{v, err} $ as follows:
 \begin{equation}
 \boldsymbol{\omega}_{des} = k_{q}\text{sign}(q_{w,err})\boldsymbol{q}_{v, err}
 \end{equation}
where $ k_{q} $ is a tuning parameter. Note that the sign of the real part of the quaternion error is used to avoid the unwinding phenomena.

The desired moments $ \boldsymbol{M}_{des} $ are computed as follows:
\begin{equation}
\boldsymbol{M}_{des} = k_{r}\left( \boldsymbol{\omega}_{des} - \boldsymbol{\hat{\omega}} \right) - \boldsymbol{r}_{off} \times {}_{B}\boldsymbol{F}_{des} + \boldsymbol{\hat{\omega}}\times \boldsymbol{J}\boldsymbol{\hat{\omega}}
\end{equation}
where $ k_{r} $ is the rate controller gain, $ \boldsymbol{\hat{\omega}} $ is the estimated angular velocity, $ \boldsymbol{r}_{off} $ is the center of mass offset, and $ \boldsymbol{J} $ is the vehicle's inertia matrix.

\subsection{Control allocation}
\label{subsec:allocation}
The control allocation problem deals with finding rotors speed $ n_{1}, \dots, n_{6} $ and rotor angular positions $\boldsymbol{\alpha}$ to satisfy equation~\eqref{eq:allocation_mapping}.
Solving this equation is challenging for two reasons.
The first reason is that any solution is not unique due of the over-actuated nature of the platform.
The second reason is that the rotors angular positions appear in a non-linear fashion in the allocation matrix~\eqref{eq:allocation_matrix}.
One approach to solve this problem would be to perform a nonlinear least square optimization. However, the computation time required and available computation resources on-board the platform inhibit such a solution.
Another approach is to use a nonlinear model predictive attitude controller \cite{kamel2015fast} that generates near optimal rotors speed and tilting angles commands. 
This approach is appealing as it can easily account for tilting angles dynamics, but could be computationally expensive.
Note that this allocation needs to be solved at high rate to allow the system to perform in an agile manner.

Here we propose an approach to transform the nonlinear allocation problem into a linear problem through variable transformation. We show that this solution also satisfies the constraint of positive rotors speed.
To this end, we decompose the force generated by each rotor as shown in equation~\eqref{eq:rotor_force_torque} along the vertical axis and the lateral axis of the rotor.
We define vertical and lateral forces of the $ i-th $ rotor as follows
\begin{align}\label{eq:variable_transformation}
	F_{v,i} &=  \mu n_{i}^2  \cos{\alpha_{i}}\\
	F_{l,i} &=  \mu n_{i}^2  \sin{\alpha_{i}}.
\end{align}
We also define the vector of dimension $ 12\times 1 $ of all vertical and lateral forces as follows
\begin{equation}
	\boldsymbol{F}_{dec} =
		\begin{pmatrix}
			F_{v,1} \\
			F_{l,1}\\
			\vdots \\
			F_{v,6} \\
			F_{l,6}
		\end{pmatrix}.
\end{equation}
By rearranging equation~\eqref{eq:allocation_mapping} we obtain
\begin{equation}\label{eq:static_allocation_mapping}
 \begin{pmatrix}
\boldsymbol{F} \\
\boldsymbol{M}
\end{pmatrix}
= \boldsymbol{A}_{\text{static}}	\boldsymbol{F}_{dec}
\end{equation}
where $ \boldsymbol{A}_{\text{static}} $ is a static allocation matrix of dimensions $ 6\times12 $ that doesn't depend on the rotors orientation $ \boldsymbol{\alpha} $.
At this point, we can easily invert equation~\eqref{eq:static_allocation_mapping} to calculate $ \boldsymbol{F}_{dec} $ from a desired wrench $ \left( \boldsymbol{F}^\top, \boldsymbol{M}^\top \right)^\top$.
We use the Moore-Penrose pseudo inverse of $ \boldsymbol{A}_{\text{static}} $ to calculate $ \boldsymbol{F}_{dec} $ as follows
\begin{equation}
 \boldsymbol{F}_{dec} = \boldsymbol{A}_{\text{static}}^{\dagger}
\begin{pmatrix}
	\boldsymbol{F}_{des} \\
	\boldsymbol{M}_{des}
\end{pmatrix}
\end{equation}

Since the system is under-determined, the Moore-Penrose pseudo-inverse is the minimum norm solution of Equation~\eqref{eq:static_allocation_mapping}. 
It is straight forward to show that the norm of $\boldsymbol{F}_{dec}$ is proportional to the sum of the fourth power of rotor speeds as shown below.
\begin{equation}
{||\boldsymbol{F}_{dec}||}^2 
= \mu^2\sum_{i=1}^6 (n_{i}^2  \cos{\alpha_{i}})^2 + (n_{i}^2  \sin{\alpha_{i}})^2 
= \mu^2\sum_{i=1}^6 n_{i}^4
\label{eq:norm}
\end{equation}
This minimization leads to more consistent and equally distributed angular velocities as well as a reduction of power consumption, which all are desirable.

Now, obtaining the actual rotor speed $ n_{i} $ and rotor orientation $ \alpha_{i} $ for each rotor is straight forward and can be done by solving the system of equations in \eqref{eq:variable_transformation}.
We have
\begin{align}
n^{2}_{i} &= \frac{1}{\mu} \sqrt{F_{v,i}^{2} + F_{l,i}^{2}}  \\
\alpha_{i} &= \text{atan2}\left( F_{l,i}, F_{v,i}\right).
\end{align}
This approach means that the calculation of motor speeds and tilting angles can be done by the simple multiplication of a matrix by a vector. 
This allows it to be calculated at a rate of several hundred Hz on the small micro-controller found in the Pixhawk.
It has to be noticed however, that allocating the actuator commands by using the pseudo-inverse may result in inadmissible rotor speeds that are higher than physically possible.
This may occur even if the desired forces and moments lie in the set of attainable forces and moments.
One approach to overcome this is the exploitation of the null space of the pseudo-inverse.
The exploration of this remains for further analysis. 

There is one major disadvantage of this allocation method, namely that the tilting angles and the angular velocities are nonlinearly coupled.
Therefore, it is difficult to constrain them separately.
This means, that this allocation doesn't account for the slower dynamics of the tilting angles.
An example of a situation in which this becomes a critical issue is hovering at a roll angle of \ang{90}. 
In this configuration one of the three rotor axes is completely vertical.
Therefore, the two thrust motors on this axis can only produce a force in horizontal direction. 
Hence, they are not used for counteracting gravity, but for disturbance rejection and horizontal position stabilization.
These two motors are required to constantly rotate \ang{180} back and forth at a high rate to provide the desired counteracting moments and forces.
The dynamics of the tilting motors are, however, much slower than that of the thrust motors and as a result can not keep up with the desired allocation commands.
This leads to forces and moments being produced in directions that have not been commanded.
To overcome this, the allocation was modified to exclude these two motors in this particular configuration.
In this configuration, the system flies with the four remaining ones, which still allow it to control all six DoF.
This is further explained in \cite{omar2017control}.


\section{Evaluation}
\label{sec:evaluation}
\subsection{Simulation}
\begin{figure}
  \centering
  \includegraphics[width=0.7\columnwidth]{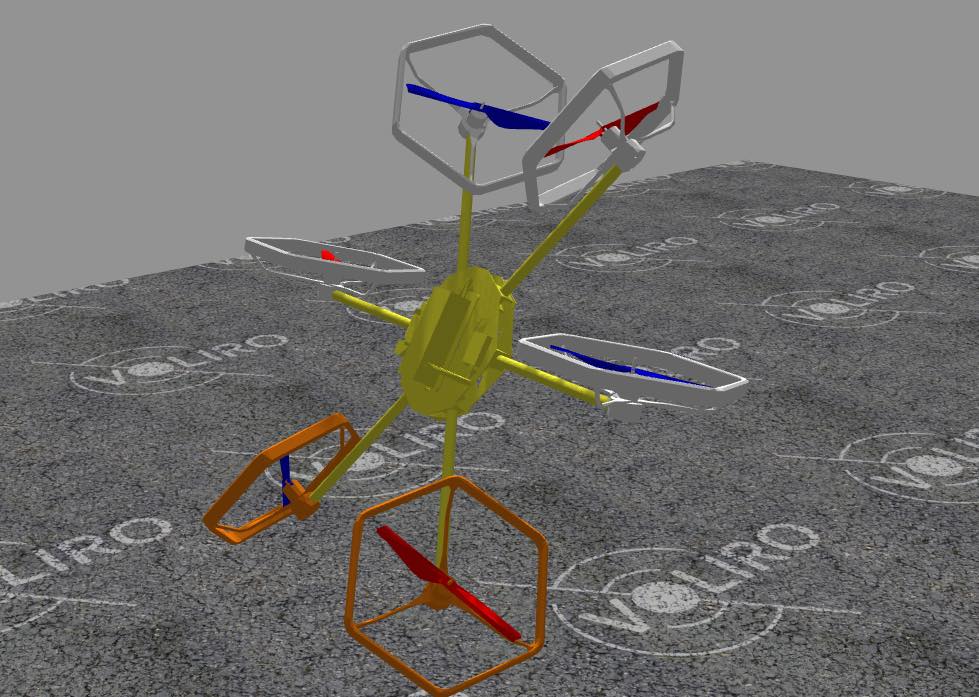}
  \caption{The model of the system in the simulation environment}
  \label{fig:simulation}
\end{figure}
The controller was tested in simulation.
This simulation uses Gazebo\cite{1389727} as a physics environment and to model the tilting motors.
To simulate the modeled sensors and the thrust motors the RotorS plugin\cite{Furrer2016} is used.
The sensor data was sent via the MAVLink protocol\footnote{pixhawk.org/dev/mavlink} to the software in the loop (SITL) version of the PX4 software.
The model of the system was then loaded into this environment.
It can be seen in Figure \ref{fig:simulation}.
Similar to the actual system, the commands are sent via ROS to the controller.

\subsection{Experimental Results}
%
%
In this section, the experimental results are presented.
The position and orientation of the vehicle are determined with an external motion capture system, which is sent to the flight controller at \SI{10}{Hz} and fused with the onboard IMU.
To showcase the vehicle’s capabilities, two different maneuvers are demonstrated.
Both maneuver are unfeasible to perform for a standard multicopter.
Moreover, we show experimental results where the Voliro platform is able to interact and move along a wall.
\subsubsection{Free Flight Experiments}

%
%
\begin{figure}
    \centering
    \begin{subfigure}[t]{0.116\textwidth}
        \centering
        \includegraphics[trim=330 10 330 110,clip,width=\textwidth]{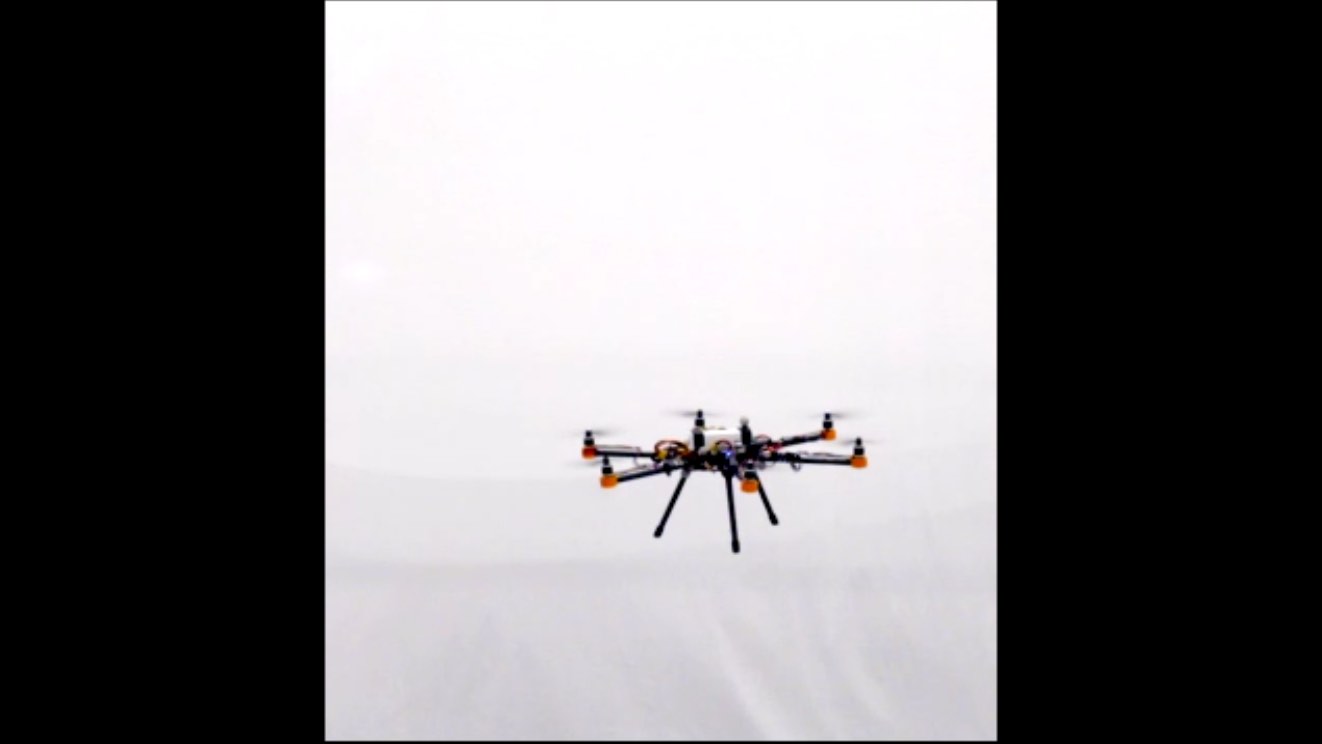}
    \end{subfigure}
    \hfill
    \begin{subfigure}[t]{0.116\textwidth}
        \centering
        \includegraphics[trim=330 10 330 110,clip,width=\textwidth]{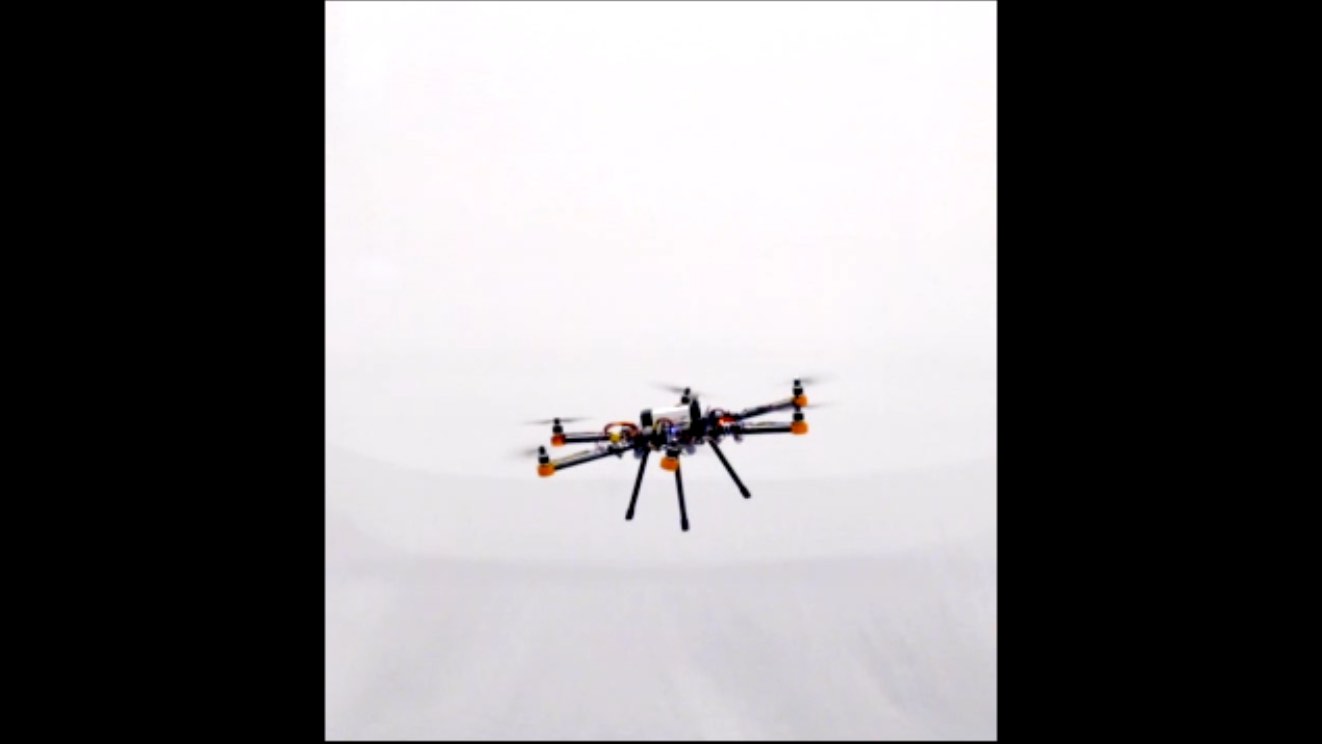}
    \end{subfigure}
        \hfill
    \begin{subfigure}[t]{0.116\textwidth}
        \centering
        \includegraphics[trim=330 10 330 110,clip,width=\textwidth]{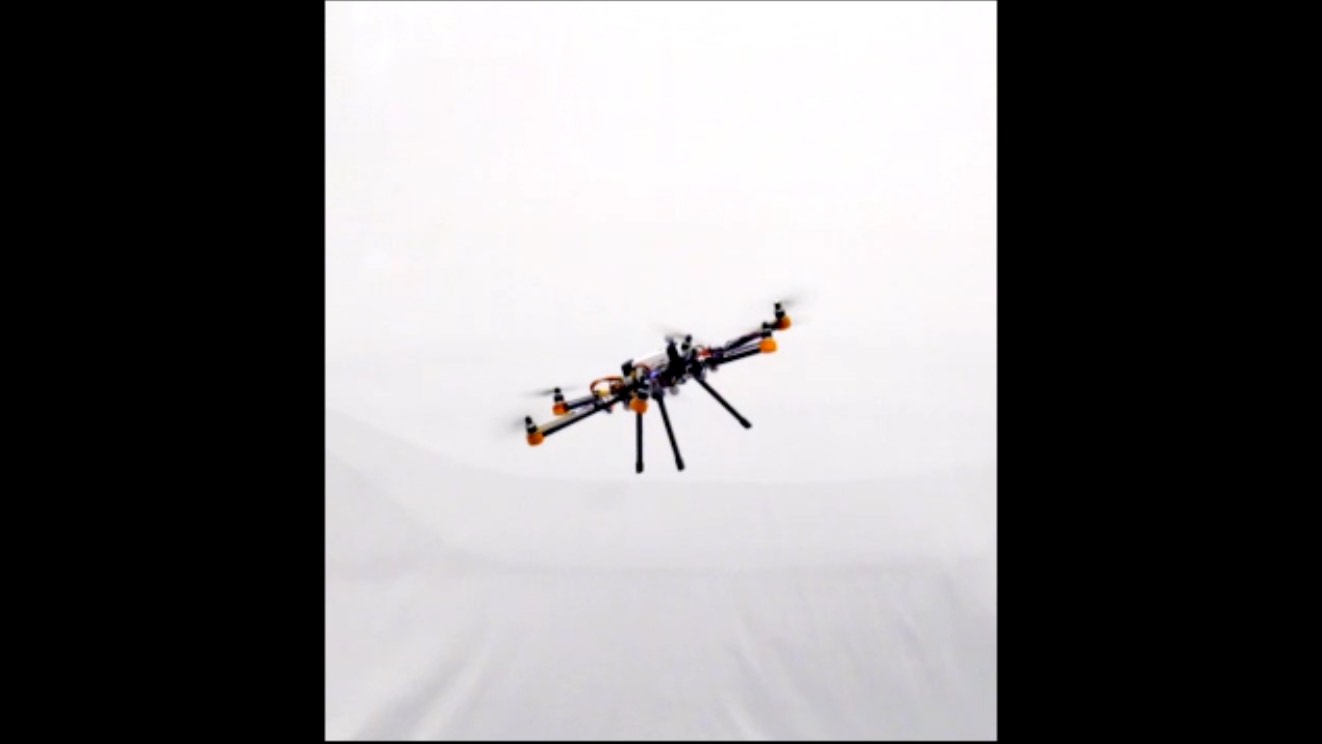}
    \end{subfigure}
        \hfill
    \begin{subfigure}[t]{0.116\textwidth}
        \centering
        \includegraphics[trim=330 10 330 110,clip,width=\textwidth]{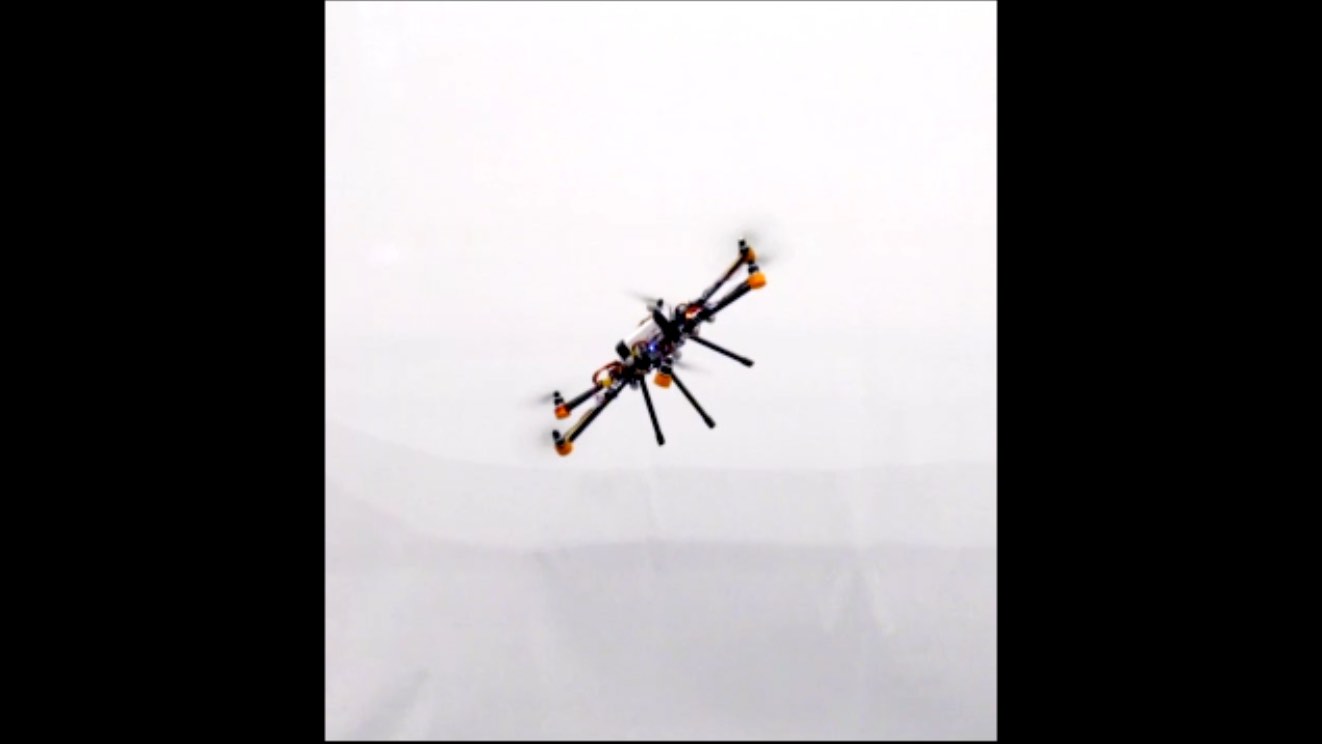}
    \end{subfigure}
        \hfill
    \begin{subfigure}[t]{0.116\textwidth}
        \centering
        \includegraphics[trim=330 10 330 110,clip,width=\textwidth]{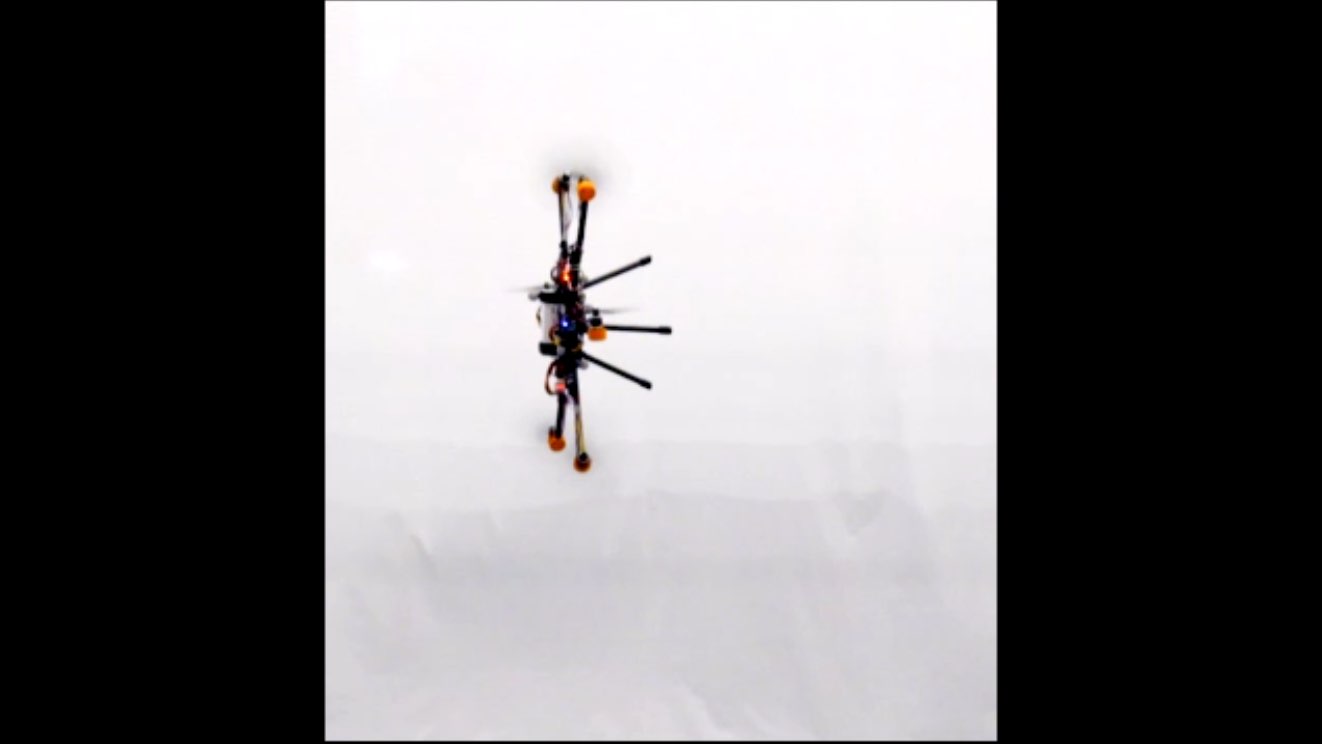}
    \end{subfigure}
            \hfill
    \begin{subfigure}[t]{0.116\textwidth}
        \centering
        \includegraphics[trim=330 10 330 110,clip,width=\textwidth]{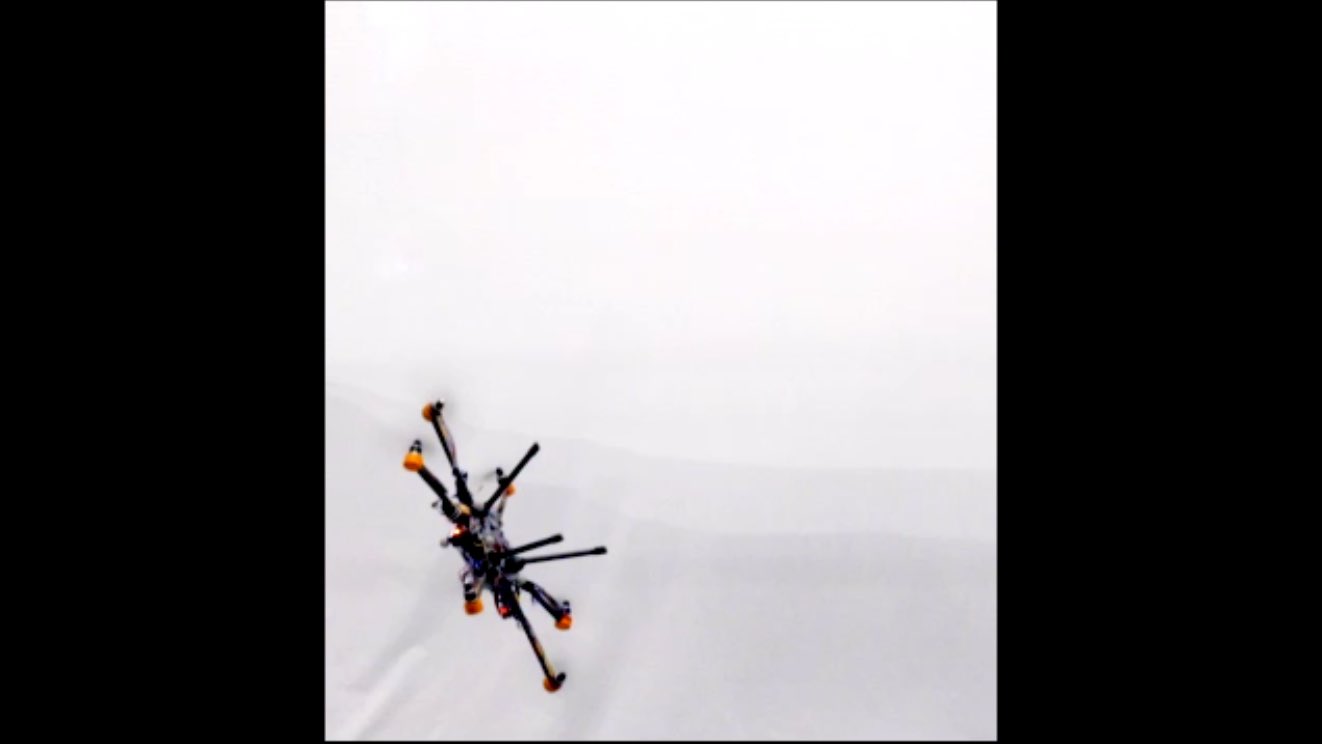}
    \end{subfigure}
            \hfill
    \begin{subfigure}[t]{0.116\textwidth}
        \centering
        \includegraphics[trim=330 10 330 110,clip,width=\textwidth]{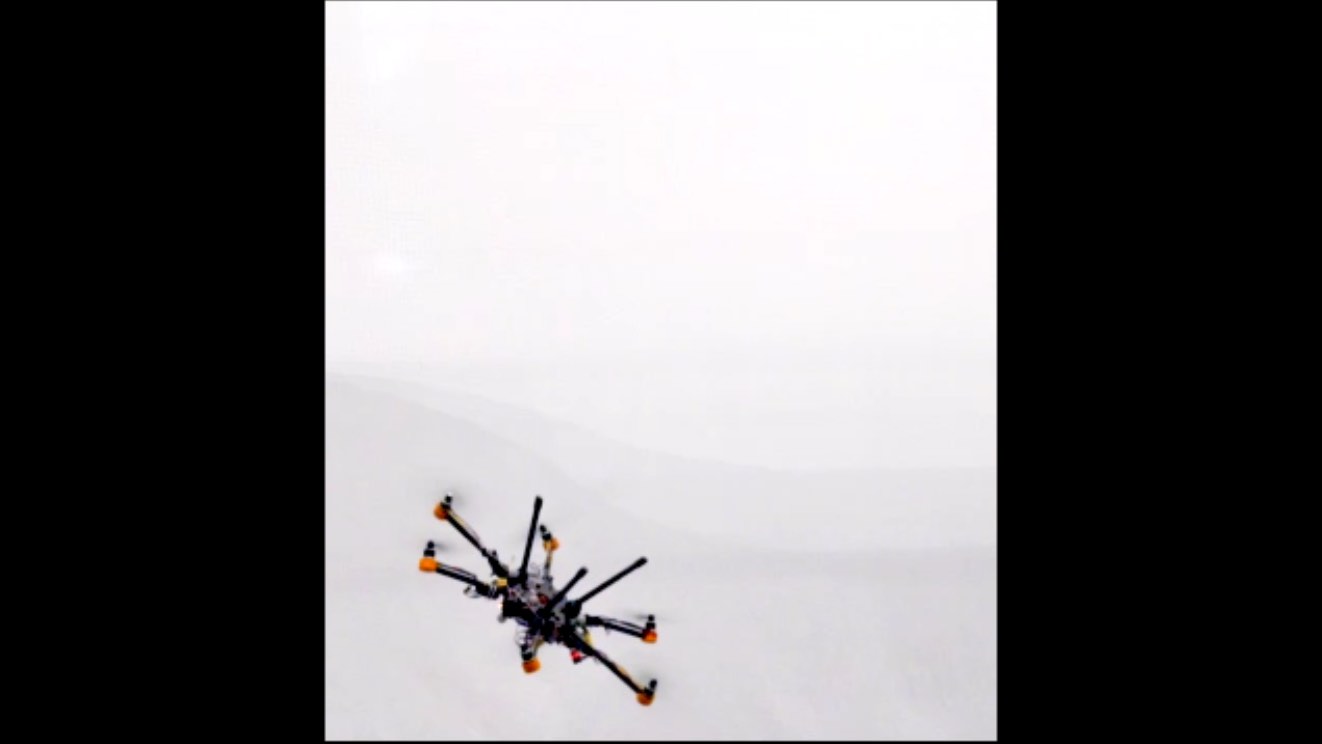}
    \end{subfigure}
            \hfill
    \begin{subfigure}[t]{0.116\textwidth}
        \centering
        \includegraphics[trim=330 10 330 110,clip,width=\textwidth]{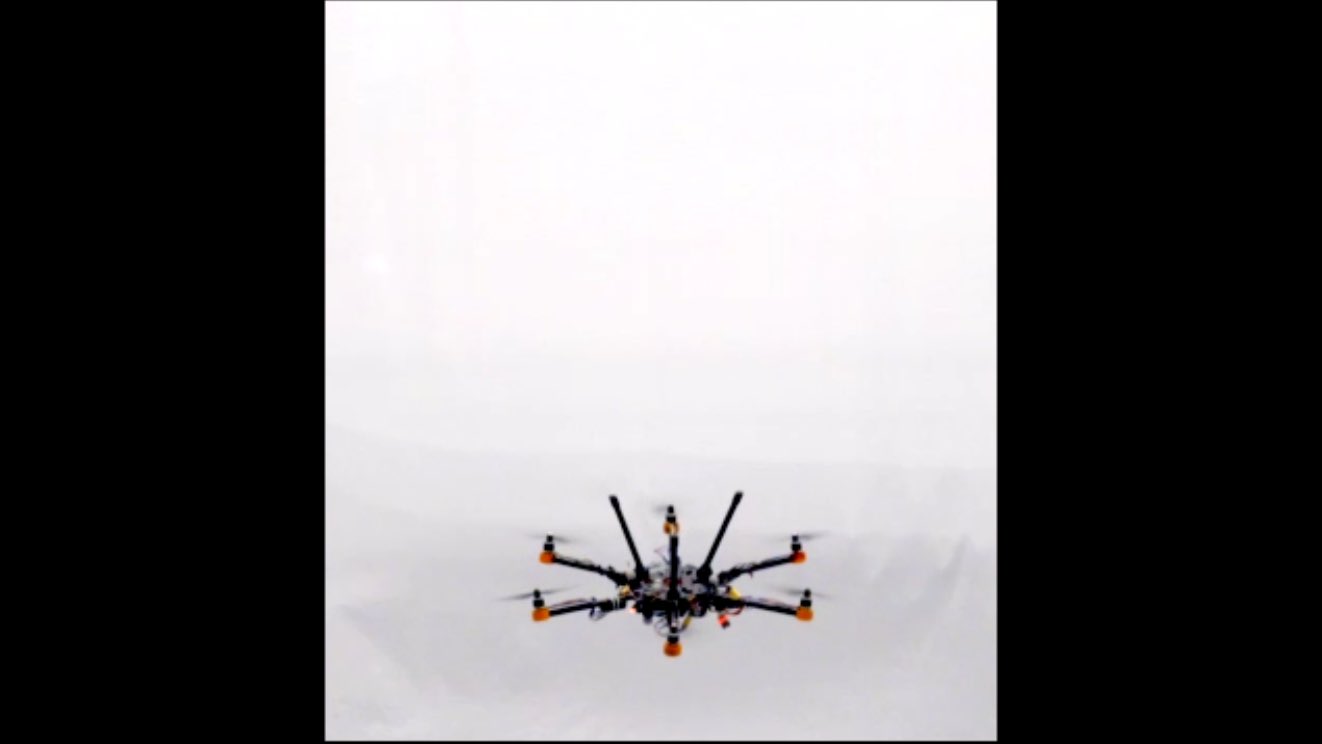}
    \end{subfigure}
    \caption{
    Snapshots of the systems rotation from horizontal to upside down flight.
    %
    %
    }
    \label{fig:upside_down_snapshots}
\end{figure}

\begin{figure}[htbp]
    \centering
        \includegraphics[width=0.45\textwidth]{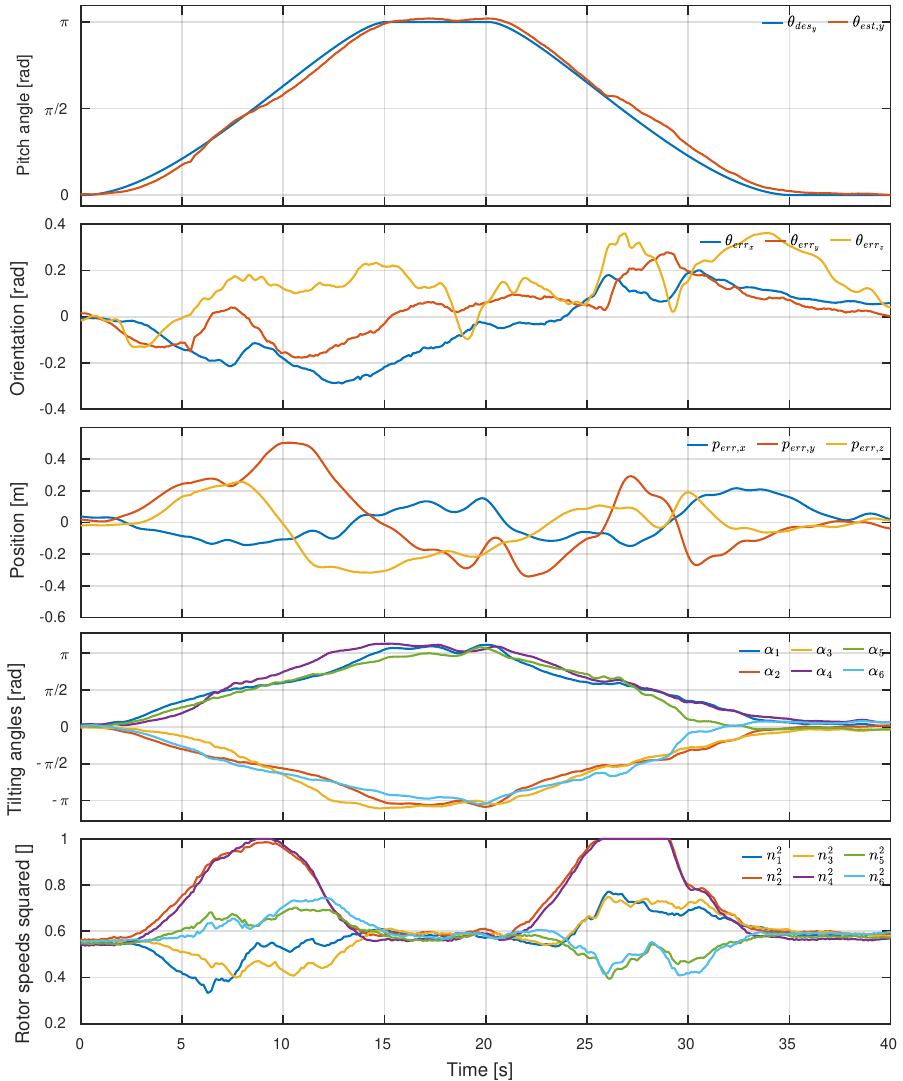}
    \caption{
    Position, orientation, tilting angles and angular velocities during the transition to upside down flight.
    The x-axis of all the plots displays the time in seconds.
    The orientation is given in zxy-Euler angles $\vec{\theta}$.
    The squared rotor speeds are scaled, where 1 represents the maximal thrust produced by a rotor.
    }
    \label{fig:upside_down_five}
\end{figure}

The first experiments show the rotation of the vehicle around the body's y-axis from its initial horizontal orientation to upside down and back. During this maneuver the other desired rotations and positions are set to zero.
The results of this experiment are shown in Figure~\ref{fig:upside_down_five}.
This maneuver is conducted in a slow manner to display the vehicle’s capability to stabilize in all the rotations around the y-axis.

\begin{figure}[htbp]
    \centering
        \includegraphics[width=0.45\textwidth]{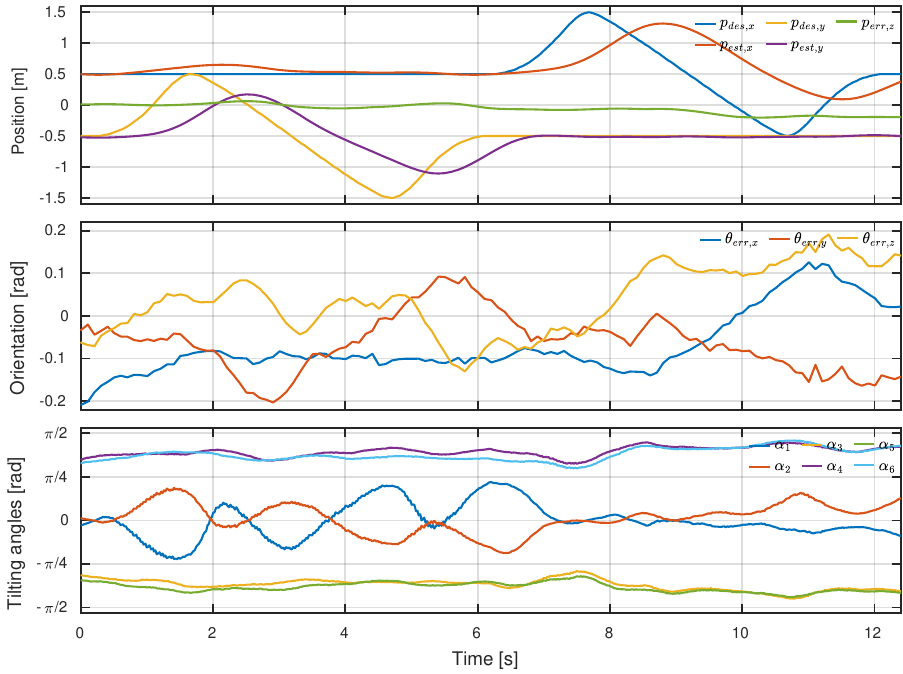}
    \caption{
    Position, orientation and tilting angle during the horizontal plus maneuver.
    The orientation is given in zxy-Euler angles $\vec{\theta}$.
    The desired rotation angle around the x-axis is \ang{50}.
    }
    \label{fig:translation_three}
\end{figure}

%
%
The second experiments displays a horizontal translation in x- and y-direction.
The desired rotation around the x-axis is set constant to \ang{50} while the other rotations are set to zero.
The results are shown in Figure~\ref{fig:translation_three}.

\subsubsection{Wall Interaction Experiments}
\begin{figure}
	\centering
	\includegraphics[width=0.4\textwidth]{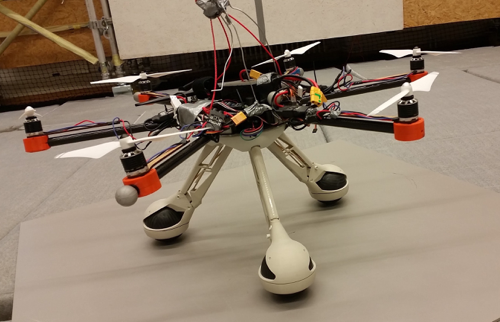}
	\caption{Three spheres module mounted on Voliro prototype.}
	\label{fig:threespheresmodule}
\end{figure}

\begin{figure}
	\centering
	\begin{subfigure}[t]{0.45\textwidth}
		\centering
		\includegraphics[width=\textwidth]{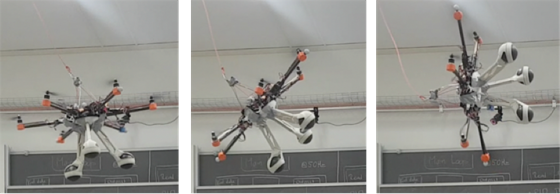}
		\caption{Voliro during a transition from horizontal to a vertical orientation.}
		\label{fig:wall_transition_to_vertical}%
	\end{subfigure}
	\vfill
	\begin{subfigure}[t]{0.45\textwidth}
		\centering
		\includegraphics[width=\textwidth]{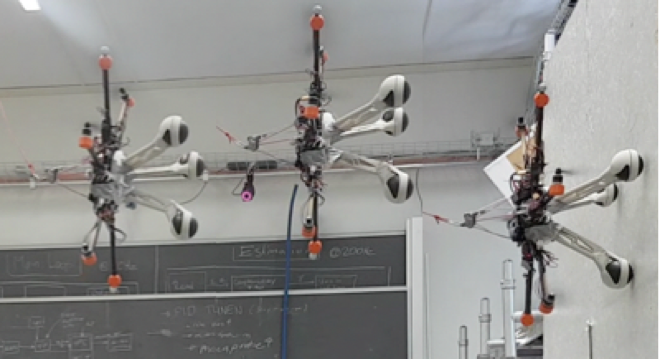}
		\caption{Approaching the wall.}
		\label{fig:wall_approaching}%
   \end{subfigure}
	\vfill
	\begin{subfigure}[t]{0.45\textwidth}
	\centering
	\includegraphics[width=\textwidth]{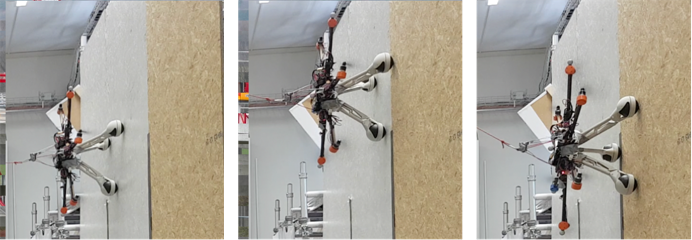}
	\caption{Driving on the wall.}
	\label{fig:wall_driving}%
\end{subfigure}
	\caption{Voliro during wall interaction.
		First, a vertical orientation is achieved to align the three spheres module with the wall (\subref{fig:wall_transition_to_vertical}).
		Then, the wall is approached while maintaining the aforementioned orientation (\subref{fig:wall_approaching}).
		Finally, the platform is able to establish stable contact with the wall and move in any direction while maintaining force against the wall (\subref{fig:wall_driving}).}
	\label{fig:wall_interaction}
\end{figure}

To enable physical interaction with the wall, a three spheres module that can roll passively is mounted on the Voliro platform as shown in Figure~\ref{fig:threespheresmodule}.
The module is passively compliant to improve  interaction stability and to reduce oscillation during contact.
The phases to achieve physical interaction with the wall can be summarized as follows:
\begin{itemize}
	\item Transition from horizontal flight to vertical flight, which is achieved at pitch  \ang{90}, as shown in Figure~\ref{fig:wall_transition_to_vertical}.
	\item Approach the wall while maintaining a pitch angle of \ang{90} until a contact is established with the compliant three spheres module. This phase is shown in Figure~\ref{fig:wall_approaching}.
	\item Driving on the wall is achieved by generating force vector using a simple proportional controller. Tracking a circle on the wall is shown in Figure~\ref{fig:wall_driving}.
\end{itemize}
More details about wall interaction control and experiments are available here \cite{paula2017development}.

The experiments demonstrate the vehicle’s omni-directionality.
It is able to obtain all orientations along one rotation axis and to perform translations at an inclined orientation.
However, the controller is not able to counteract all of the system’s translational and rotational dynamics, such that the position and the orientation are still slightly coupled.
This can be partially explained by the slower dynamics of the tiling motors when compared to the thrust motors.
The experiments showed that the roll and pitch angles are better tracked than the position. This is because the rotation dynamics is controlled mainly by changing the rotational speed of the rotors and utilizes their fast dynamics.

Deviations in position and in yaw are generally corrected by the tilting of the rotors and are more sensitive to their slow dynamics.

While flying at a pitch angle \ang{90}, the allocation demands for inadmissible rotor speeds, leading to two thrust motors becoming saturated.
This is further discussed in section \ref{subsec:allocation}.
Nevertheless, the system is still able to track the desired position and orientation.

\section{Discussion and Conclusion}\label{sec:conclusion}
This work has presented the Voliro platform, a hexacopter with tiltable rotors. 
We demonstrated how this basic idea can be used to achieve omnidirectional maneuverability while avoiding wasting energy on generation of counteracting  forces which is an issue that inhibits the use of fixed orientation rotors in omni-directional designs. 
We presented the mechanical design of the platform and a compact tiltable rotor.
We also presented a control scheme with an innovative control allocation technique. 
The main advantage of the allocation technique presented in this work is the simplicity and the limited computation effort required to computed tilting angles and rotor speed. 
However, problems caused by slow rotor tilting dynamics remain an open issue.
In various experiments, we demonstrated a transition from horizontal to upside flight and physical interaction with a wall.
While the increased maneuverability gives rise to a broader scope of applications, some of them may require a powering tether in order to overcome the limitations imposed by battery life.
In conclusion, tiltable rotor multirotor showed great capabilities and can push the boundaries of what is currently achievable by standard fixed orientation multirotors.

\section*{Acknowledgment}
The authors would like to acknowledge Philipp Andermatt, Cliff Li. Alexis Müller, Kamil Ritz, and Kevin Schneider who were part of the Voliro focus project team and contributed in numerous ways to the development and deployment of the Voliro prototypes.

\bibliographystyle{IEEEtran}
\bibliography{bibliography}
\end{document}